\documentclass{article}
\usepackage{ijcai20}

\usepackage{times}
\usepackage{soul}
\usepackage{url}
\usepackage[hidelinks]{hyperref}
\usepackage[utf8]{inputenc}
\usepackage[footnotesize]{caption}
\usepackage{graphicx}
\usepackage{amsmath}
\usepackage{amsthm}
\usepackage{booktabs}
\usepackage{caption}

\usepackage{algorithm}
\usepackage{algorithmic}
\usepackage{soul}

\usepackage[dvipsnames]{xcolor}

\title{ChemoVerse: Manifold traversal of latent spaces for novel molecule discovery}

\author{
Harshdeep Singh$^1$\footnote{This work was performed during an internship programme at Accenture Labs, Dublin, Ireland.}
\and
Nicholas McCarthy$^2$\and
Qurrat Ul Ain$^3$\and
Jeremiah Hayes$^2$ 
\affiliations
$^1$Swiss Federal Institute of Technology, Lausanne, 
$^2$Accenture Labs, Dublin
$^3$AI Innovation Lab, DSAI, Novartis
\emails
harshdeep.harshdeep@epfl.ch,
\{nicholas.mccarthy,jer.hayes\}@accenture.com,
qurrat.ul\_ain@novartis.com
}

\begin{document}

\maketitle

\begin{abstract}
 In order to design a more potent and effective chemical entity, it is essential to identify molecular structures with the desired chemical properties. Recent advances in generative models using neural networks and machine learning are being widely used by many emerging startups and researchers in this domain to design virtual libraries of drug-like compounds. Although these models can help a scientist to produce novel molecular structures rapidly, the challenge still exists in the intelligent exploration of the latent spaces of generative models, thereby reducing the randomness in the generative procedure. In this work we present a manifold traversal with heuristic search to explore the latent chemical space. Different heuristics and scores such as the Tanimoto coefficient, synthetic accessibility, binding activity, and QED drug-likeness can be incorporated to increase the validity and proximity for desired molecular properties of the generated molecules. For evaluating the manifold traversal exploration, we produce the latent chemical space using various generative models such as grammar variational autoencoders (with and without attention) as they deal with the randomized generation and validity of compounds. With this novel traversal method, we are able to find more unseen compounds and more specific regions to mine in the latent space. Finally, these components are brought together in a simple platform allowing users to perform search, visualization and selection of novel generated compounds.
 

\end{abstract}

\section{Introduction}

Designing a new chemical entity is a time consuming, expensive, and error-prone task. Pharmaceutical companies invest billions of dollars into screening vast libraries of chemical compounds for hit and lead identification \cite{NatureBillions}. The past few years has seen the rise of deep generative models that can operate over large spaces of molecular structures and embed the chemical properties of such into a vector space. By decoding from this 'latent' space of chemical structure we can generate new, previously unidentified chemical compounds.


In the domain of computational chemistry, Simplified molecular-input line-entry system or SMILES are a common string textual method for encoding and representation of molecular structures \cite{anderson1987smiles}. This facilitates the use of models more commonly used in natural language processing. Therefore, SMILES strings have been used as raw input strings to generative models, which are given the task of encoding and decoding the SMILES string directly \cite{anderson1987smiles}. Advances were made by employing variational autoencoders (VAE), a neural network comprised of an encoder that transforms a compound's representation into a compressed latent space, and a decoder that generates compounds from the latent space \cite{kingma2013auto}. Although, directed search of the resulting latent space is difficult. To counter this, conditional variational autoencoders (CVAE)\cite{kang2018conditional} were used in order to facilitate the generation of new molecules with specified molecular properties. This is achieved by incorporating the molecular properties of a compound into the encoder layer and helping in the generation of more drug-like molecules \cite{lim2018molecular}. Generative adversarial networks (GANs) have also been applied in the same manner \cite{maziarka2020mol}, and have been recently combined with reinforcement learning and graph representation of molecules to optimize the generation of molecules with specified molecular properties~\cite{de2018molgan}.

Discovery in the latent space generated by these models is often performed using random sampling and linear interpolation, primarily due to the ease of the implementation of these methods. However, this is not suitable for most generative models as their latent spaces are generally high dimensional and sparse. While doing traversal, we will traverse regions where the data is not very well represented. In other words, this could lead to a 'dead zone' as the space of molecular samples in the training dataset are present only on a subset of the latent space \cite{white2016sampling}. Hence, decoding a point from the latent space will end up returning noisy or invalid results. It can also be challenging to incorporate contextual domain information during search, and as a result discovery of compounds with specific properties is often very inconsistent. 

\begin{figure*}[t]
\begin{center}
\begin{minipage}[c]{0.715\linewidth}
\includegraphics[width=\linewidth]{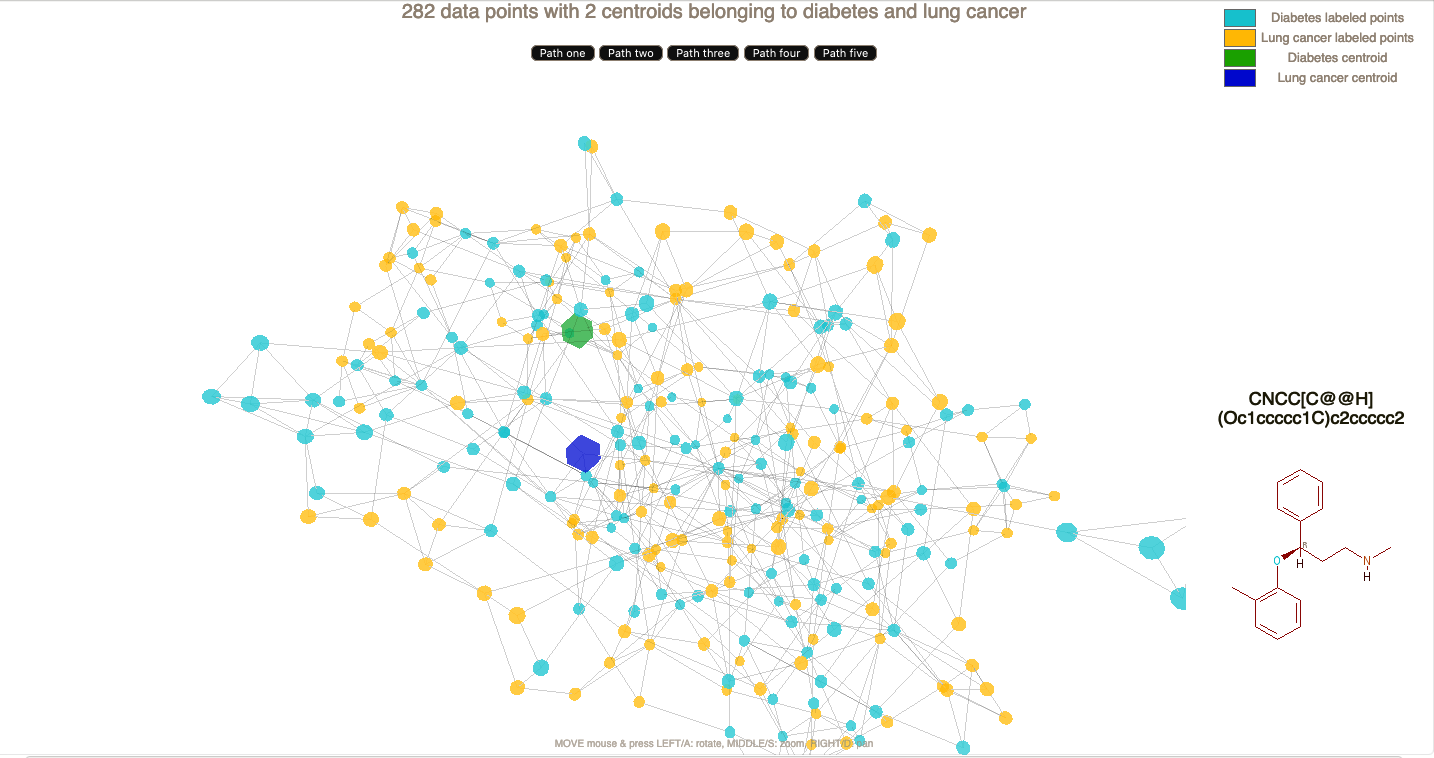}
\caption{Visualization of the latent space as a K-dimensional tree, while traversing the latent manifold of compounds used in the treatment of diabetes and lung cancer.}
\label{fig:fig_screenshot_one}
\end{minipage}
\end{center}
\end{figure*}

\begin{figure*}[t]
\begin{center}
\begin{minipage}[t]{0.715\linewidth}
\includegraphics[width=\linewidth]{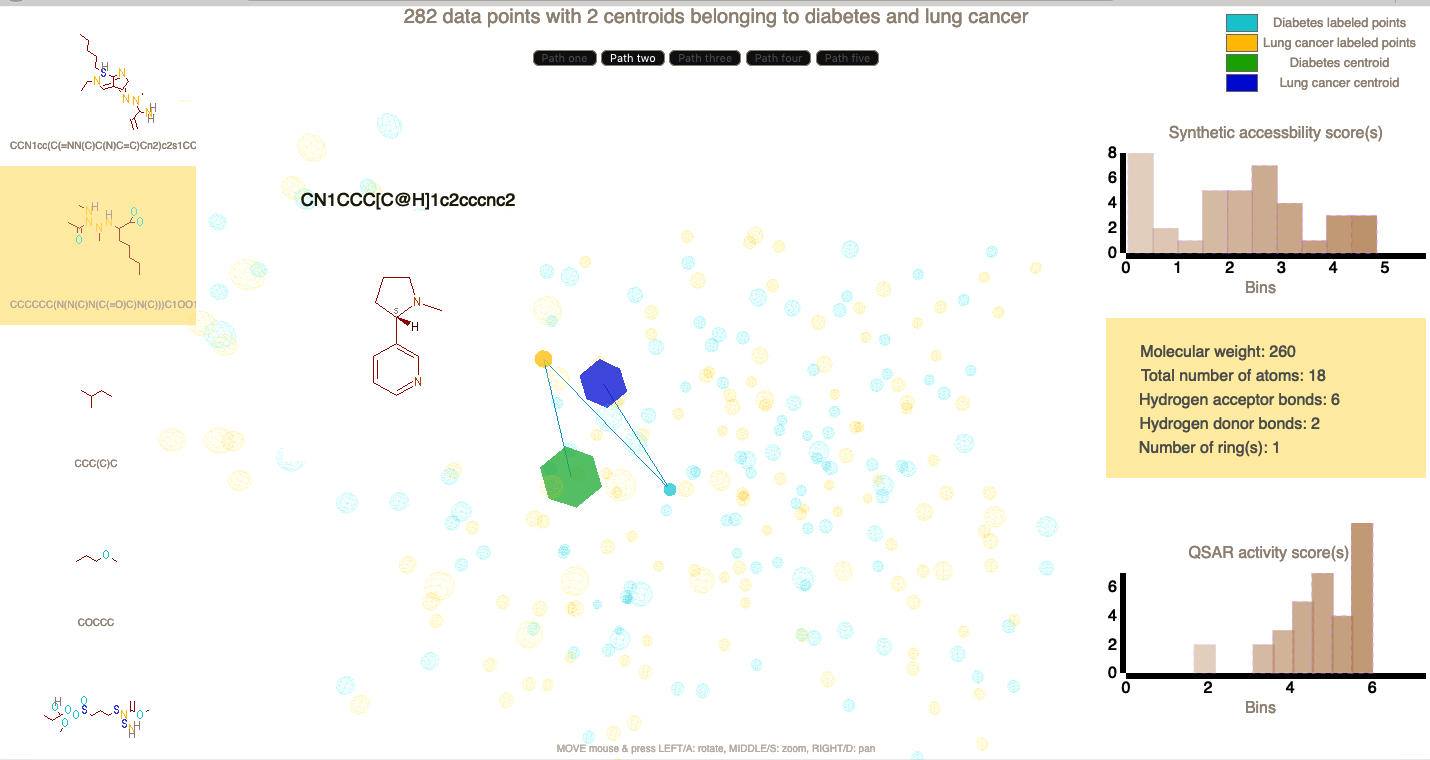}
\caption{Visualization of the path in the latent space along which our algorithm is interpolating. On the left are generated compounds not present in the training data. On the right are histograms of different molecular properties, including synthetic accessibility, activity scores.}
\label{fig:fig_screenshot_two}
\end{minipage}%
\end{center}
\end{figure*}

In this work we implemented various flavours of auto-encoders as the generative model for producing sets of latent spaces, and in particular show that our novel implementation of Grammar VAE \cite{kusner2017grammar} with an additional attention mechanism \cite{vaswani2017attention} is highly performative, with a low rate of invalid molecules generated. We also introduce a novel manifold interpolation method employing the Riemannian metric~\cite{arvanitidis2017latent} in conjunction with a set of molecular property heuristics to perform directed search and interpolation of these latent spaces in order to design novel molecules with desired properties. This combination of generation and exploration of latent space has enabled us to not only design molecules which have not been seen before but also to explore new regions of latent chemical space where more potent chemical compounds may exist.


\section{System Architecture}

In this section we describe the various components of our system architecture: generation of latent spaces and our algorithm for manifold traversal.



\subsection{Data}

We used a dataset of 250,000 molecules drawn from the ZINC dataset \cite{irwin2012zinc}, and an additional 100,000 drawn from the ChEMBL dataset~\cite{gaulton2017chembl}. These two datasets are comprised of commercially available drug molecules and have been used in related work using models like variational autoencoders (VAEs)~\cite{gomez2018automatic}. Molecules are represented in canonical SMILES string format, and are further processed into 1)~a one-hot character encoding and 2)~a set of context-free grammar (CFG) rules. Grammar rules are obtained from the OpenSMILES specification~\cite{opensmiles}, which denotes how the SMILES representation was formed based on the rules. This context free grammar (CFG) consists of 76 production rules, to which an additional seven are added, and a further nine modified in order to represent the more complex ChEMBL dataset.



\subsection{Latent Space Generation}

Three models are implemented in our system: a VAE \cite{kingma2013auto}, a Grammar VAE \cite{kusner2017grammar}, and a Grammar VAE with self-attention \cite{vaswani2017attention}. As our search algorithm is model agnostic, latent spaces can be substituted with minimal effort. However, results will vary depending on the underlying data, model architecture, and training parameters used to generate each latent space. 


The ChEMBL dataset is less standardized and contains more complex molecules, therefore we perform transfer learning by initially training each model on the ZINC dataset for 50 epochs, then switching to the ChEMBL dataset for 50 epochs. Training, validation and test sets of 85\%, 10\%, and 5\% respectively was used, with the test set comprised entirely of ChEMBL molecules. We use the Adam optimizer with a learning rate scheduler that is instantiated after 15 epochs with a factor of 0.1 (initialized at 0.001). 

The encoder is comprised of three 1D convolutional layers with filters of size 9, 10 and 11 respectively, while the decoder is comprised of 3 gated recurrent units (GRU) of 501 units~\cite{kusner2017grammar}. The structural validity of the generated compounds are checked using the open-source RDKit library \cite{rdkit}. Examples of test set compounds that have been encoded and decoded are shown in table~\ref{tab:generated_compounds}. 


As noted in the original Grammar VAE work \cite{kusner2017grammar} the vanilla VAE architectures encoding SMILES strings directly  generally produce a very low valid decode rate on larger molecular datasets - just 17\% using a conditional VAE under their bayesian optimization search methodology. By instead using a Grammar VAE to generate production rules of a grammar instead of SMILES strings directly a much higher rate of valid compounds was attained. However, as OpenSMILES is a context-free instead of regular grammar it is still unable to model certain subtle characteristics of the SMILES grammar such as paired ring bonds; for example the SMILES string 'c1ccccc1C2CCCC2' would be decoded as 'c1ccccc1C2CCCC', incorrectly dropping the final paired digit. By incorporating a self-attention layer in the Grammar VAE architecture this effect was mitigated, and increased the validity of decoded test set molecules from 61\% to 70\%. 

\begin{table*}[ht]
	\caption{Input and corresponding decoded SMILES strings of test set compounds generated by the model. Tanimoto similarity is used to measure the the structural similarity. An observation that can be made here is that small changes in the generated structure w.r.t to actual structure might lead to disproportionate changes in the similarity.}
	\centering
	\begin{tabular}{p{0.42\linewidth}p{0.42\linewidth}p{0.1\linewidth}}
		\hline
		\textbf{Actual Compound} & \textbf{Generated compound} & \textbf{Similarity} \\
		\hline
		OC(=O)CSc1oc2ccccc2n1 & OC(=O)CCS1cc2ccccs2n1 & 0.141 \\
		CC(=O)C(=Cc1ccccc1)c2ccccc2 & O.CCCN(Cc1ccccc1c2)cccnn2 & 0.171 \\
		Cc1occc1C(=O)NCc2onc(n2)c3ccccc3 & Nc1nnc1NC(=O)NCSc2nn(s2)c3ccccc3 & 0.221 \\
		Cc1ccc(SCC(=O)NNC(=O)c2cncc(Br)c2)cc1 & Clc1ccc(SCC(=O)NC(=O)c2cncc(Br)c2cc1)CF & 0.373 \\
		CCc1ccc(cc1)N(CC(=O)N2CCOCC2)S(=O)(=O)C & CSc1ccc(cc1N(C(=O)Oc2CCCCC2S(=O)(=O)))ON & 0.235 \\
		\hline
	\end{tabular}
	\label{tab:generated_compounds}
\end{table*}

\begin{table*}[ht]
    	\caption{Samples of the compounds generated while interpolating on the latent space path as shown in Figure~\ref{fig:fig_screenshot_two}. These compounds are not present in the train/test datasets, and we assign a potential target label based on labels of their nearest neighbours in the test set. SAS score indicates if a molecule is difficult to synthesize. According to the Lipinski's rule of five~\protect\cite{benet2016bddcs}, the molecular mass for an active drug  should be less than 500 daltons. Activity range demonstrates the binding activity of a compound: the greater the range more active it is.}
	\centering
	\begin{tabular}{p{0.42\linewidth}p{0.14\linewidth}p{0.07\linewidth}p{0.09\linewidth}p{0.14\linewidth}}
		\hline
		\textbf{Generated compound} & \textbf{Activity range} & \textbf{SAS Score} & \textbf{Molecular weight} & \textbf{Potential Label}\\ 
		\hline
		CS1(C2)CC(C(C3C(C))C)[C@]1n1C2n1C3C4CC4C & Less than 5 & 6.186 & 299.89 & DIABETES \\
		CCC1(C)C(C)CCCCS1C(C1(O([C@+2](C1))))S & Between 5 and 7 & 6.251 & 274.49 & DIABETES \\
		CSCCCC2(C=C(O))S(C=O)C2C1[C@@](C1(C))CC & Between 5 and 7 & 5.990 & 301.49 & DIABETES \\
		CC1(N(S))NN1CCCNc1ncnc2CCn2c1CC & Greater than 7 & 5.233 &  299.89 & LUNG CANCER\\
		CS1(C2)CC(C(C3C(C))C)[C@]1n1C2n1C3C4CC4Cl & Less than 5 & 6.186 & 309.44 & LUNG CANCER\\
		CCCC1CCC(S)N(C1)C(O)NCC	& Less than 5 & 4.355 & 232.93 & DIABETES\\
		\hline
	\end{tabular}
	\label{tab:newly_found_compounds_latent_space}
\end{table*}

\begin{algorithm}[tb]
  \caption{Interpolation with manifold traversal}
  \label{alg:riemann_algo}
  \begin{algorithmic}
    \STATE \textbf{Input}: latent space $L$, source and destination points $s$ and $d$, points of interest $R$, path length $m$
    \STATE \textbf{Parameter}: $k$ heuristics and associated edge weights $W^k$
    \STATE \textbf{Output}: a path of $m$ equidistant points from $s$ to $d$
    \FOR{each reference point \textit{i} in $R$}
      \STATE Find \textit{n} nearest neighbors of $i$ $\rightarrow$ $N_i$
      \FOR{each (\textit{i}, \textit{j}), where \textit{j} $\in$ $N_i$}
        \STATE 1. Calculate Jacobian matrix $J_{ij}$ of decoder output ($dec$) w.r.t to latent space ($L$) $\frac{\partial dec(L_ij)}{\partial L_{ij}}$
        \STATE 2. Obtain $k$ heuristic distances $H_{ij}^k$ between the 2 points
        \STATE 3. Determine path edge weight based on Jacobian and heuristic:  $P_{ij}$ $\leftarrow$ $\sum_{ij}^{N} J_{ij}$ + $\sum_{k} H^k_{ij}$ $\cdot$ $W^k$ and store in a k-d tree.
      \ENDFOR
    \ENDFOR
    \STATE Apply a path finding algorithm such as Yen's algorithm (with A*) to determine a path from $s$ to $d$ on the k-d tree. 
    \STATE Segment the path using $m$ equidistant points.
    \STATE \textbf{return} $P$ $\leftarrow$ ($s, p^1, .. , p^{m-2}, d $)
  \end{algorithmic}
\end{algorithm}

\subsection{Manifold Traversal of Latent Space}

Once a latent space has been generated we can apply our manifold traversal algorithm to generate interpolative paths in the latent space in order to decode molecules with desired properties. A common approach here is to use linear or spherical interpolation~\cite{white2016sampling}, however both approaches assume that the latent space is Euclidean and flattened out, and generally produce noisier results~\cite{arvanitidis2017latent}. In our algorithm, source and destination points are selected in the latent space; these can either be latent space encoding of single molecules, or cluster centroids of molecules labeled with a desired property. Points of interest are the set of known molecules with desired properties, for example all molecules used in treatment of a specific condition. The goal is to define a path from source to destination points in the latent space through regions of interest, factoring in any additional user-specified heuristics such as synthetic accessibility, binding activity, or drug-likeness that will augment generated molecules. 

Interpolation is performed by first calculating the Jacobian distances for all points of interest. This helps us to understand how much each latent space point differs from another based on the representation learnt by the model, and understand the stretching and rotational transformations of the local neighborhood of each point with respect to other points. 

A k-dimensional tree is built using the resulting Jacobian distances as edge weights between compounds in the points of interest. Therefore, placing compounds with greater structural similarity in closer proximity on the tree. A k-d tree is chosen since it divides the domain of search into half at each level. Hence search for a node in the tree can be done in logarithmic time and makes the data structure run time efficient. Edges can also be weighted by user-specified domain heuristics, adding a weighted cost to augment the paths produced to generate molecules more relevant for a specific target. These heuristics are listed below: 

\begin{itemize}
    \item Fingerprint Similarity: A fingerprint is a series of binary digits (bits) that represent the presence or absence of particular substructures in the molecule. The similarity can be tested using cosine or Tanimoto distance metrics. The Tanimoto similarity takes into account the structural properties whereas cosine does not.
    \item Synthetic Accessibility (SA Score): A molecule synthetic accessibility is a score which is between 1 (easy to produce) and 10 (very difficult to produce). This is calculated based on fragment contributions and molecule complexity. The absolute difference between two molecules is taken into account.
    \item Drug-likeliness: This is a score which takes into account if the molecule is 'drug-like'. This is evaluated using several parameters such as molecular weight, solubility in water or lipophilic efficiency. The absolute difference between two molecules is taken into account.
    \item{Binding activity}: This demonstrates the potency to a target for a potential drug compound; less than 5 is considered inactive, 5-7 of intermediate activity, greater than 7 active.
\end{itemize}

Yen's algorithm \cite{yen1971finding} in combination with the A* algorithm is then applied on the k-d tree to find the shortest path from source to destination given the user constraints. Once the shortest path is found, we interpolate along this path equidistantly and decode the points on the latent space using the generator to generate compounds. Multiple paths can be found by taking into account the shortest path and either perturbing it or by changing the number of interpolation points between the source and the destination, and intuitively this increases the overall number of novel generated compounds. 

Manifold traversal is inherently more useful than linear or spherical interpolation as it gives users greater flexibility in path exploration under various conditions, and empirically demonstrates a much higher rate of valid decoded molecules. For example, when considering the \textit{diabetes} and \textit{lung cancer} centroids; linear interpolation with 100 equidistant points decoded along the path of centroids generated just 3 compounds with valid structures. On the contrary, applying manifold traversal with fingerprint similarity and Yen's algorithm as heuristic and perturbing the source and destination points produced 4 different paths. These four paths generated a total of 156 valid, novel compounds along the interpolated manifold between the latent regions of \textit{diabetes} and \textit{lung cancer} labeled molecules. Samples of these generated compounds can be seen in Table \ref{tab:newly_found_compounds_latent_space}. Specific regions to mine within the latent space can be found by plotting different paths, bound them and finding the overlap region where compounds with the right structure and specific characteristics can be found. 



\section{Conclusions and Future Work}
In this work we presented a model-agnostic platform for performing manifold traversal on latent spaces with user specified domain heuristics. This interpolation method allows us to add more context and direction to search and discovery of molecules in the latent space. Methods for exploration of latent spaces generated from datasets of millions of molecules provide an extremely valuable tool for virtual drug screening, and an ability to facilitate rapid drug discovery. 

Some avenues for future work in this domain include: implementation of alternative models to produce latent-spaces of various characteristics; more sophisticated methods for curve fitting in high dimensional spaces such as Bézier curves and Gaussian regression; alternative search methods such as using evolutionary and genetic algorithms on the latent space. Future work would also focus on implementing latent space evaluation metrics using this interpolation method to understand the underlying aspects of these spaces.

\appendix


\clearpage
\bibliographystyle{named}
\bibliography{ijcai20}

\end{document}